\documentclass{article}

\usepackage{graphicx}
\usepackage{subfigure}
\usepackage{booktabs} 

\usepackage{hyperref}
\usepackage{times}
\usepackage{helvet}
\usepackage{courier}
\usepackage{url}
\usepackage{amsmath}
\usepackage{amssymb}
\usepackage{amsthm}
\usepackage{amsfonts}
\usepackage{bm}
\usepackage{multirow}
\usepackage{algorithm,algorithmic}
\usepackage{tcolorbox}
\usepackage{epstopdf}

\usepackage[accepted]{sysml2019}

\def\argmin{\mathop{\hbox{argmin}}\limits}

\def\floor#1{{\left\lfloor\,#1\,\right\rfloor}}

\newcommand{\setlinespacing}[1]%
           {\setlength{\baselineskip}{#1 \defbaselineskip}}

\newcommand{\sabs }[1]{|#1|}

\sysmltitlerunning{K-TanH:  Efficient TanH for Deep Learning}

\begin{document}

\twocolumn[
\sysmltitle{K-TanH: Efficient TanH for Deep Learning}



\sysmlsetsymbol{equal}{*}

\begin{sysmlauthorlist}
\sysmlauthor{Abhisek Kundu}{pcli}
\sysmlauthor{Alex Heinecke}{pcls}
\sysmlauthor{Dhiraj Kalamkar}{pcli}
\sysmlauthor{Sudarshan Srinivasan}{pcli}
\sysmlauthor{Eric C. Qin}{gatech}
\sysmlauthor{Naveen K. Mellempudi}{pcli}
\sysmlauthor{Dipankar Das}{pcli}
\sysmlauthor{Kunal Banerjee}{pcli}
\sysmlauthor{Bharat Kaul}{pcli}
\sysmlauthor{Pradeep Dubey}{pcls}
\end{sysmlauthorlist}

\sysmlaffiliation{pcli}{Parallel Computing Labs, Intel Labs, Bangalore, India}
\sysmlaffiliation{pcls}{Parallel Computing Labs, Intel Labs, Santa Clara, USA}
\sysmlaffiliation{gatech}{Georgia Institute of Technology, USA}

\sysmlcorrespondingauthor{Abhisek Kundu}{abhisekkundu@gmail.com}

\sysmlkeywords{Approximate Tanh, Deep Learning}

\vskip 0.3in

\begin{abstract}
We propose K-TanH, a novel, highly accurate, hardware efficient approximation of popular activation function TanH for Deep Learning. 
K-TanH consists of parameterized low-precision integer operations, such as, shift and add/subtract (\textit{no floating point operation needed}) where parameters are stored in  very small look-up tables that can fit in CPU registers. K-TanH can work on various numerical formats, such as, Float32 and BFloat16. High quality approximations to other activation functions, e.g., Sigmoid, Swish and GELU, can be derived from K-TanH.  
Our AVX512 implementation of K-TanH demonstrates $>5\times$ speed up over Intel SVML, and it is consistently superior in  efficiency over other approximations that use floating point arithmetic. Finally, 
we achieve state-of-the-art Bleu score and convergence results for training language translation model GNMT on WMT16 data sets with approximate TanH  obtained via K-TanH on BFloat16 inputs.
\end{abstract}

]



\printAffiliationsAndNotice{ }  

\section{Introduction}\label{sec:intro}

Most of the compute in current Deep Learning workloads is General Matrix Multiplication (GEMM) operations, therefore the trend of efficient DL research is to optimize the GEMM kernel through software and/or hardware accelerators (see \cite{AccelatorSurvey} for survey). Non-GEMM operations are dominated by computation of activations which are critical for non-linear representation ability of neural networks while they perform complex tasks, such as, image classifications and language translations. Popular choices of activations are TanH and Sigmoid for language translations, and ReLU \cite{relu} for image classifications. 
Very recently, Swish \cite{swish17} and Gaussian Error Linear Units (GELUs) \cite{gelu} are shown to achieve higher accuracy than  ReLU for image classification and NLP and speech tasks. 
%
Exact computation of these functions are expensive operations as they involve computation of exponential function. %
GEMM ops can be made efficient with low-precision kernels, e.g., 16-bit, 8-bit arithmetic operations,  with float32 accumulator. Float32 numbers can be quantized to 16-bit formats before sending to lower memory in order to reduce data movement cost. 
With such acceleration of GEMM ops for data centers and extreme low-precision binary/ternary inference on edge devices, the percentage of time spent computing such activations will become more significant. 

For input $x$, TanH is defined as :
$$
\text{TanH}(x) = \frac{e^x - e^{-x}}{e^x+e^{-x}} = 1 - \frac{2}{1+e^{2x}} \in [-1, 1]
$$
Sigmoid can be derived from TanH as
$$
\text{Sigmoid}(x) = (1 + \text{TanH}(x/2)/2
$$ 
Similarly, Swish and GELU can be implemented using TanH.
$$
\text{Swish}(x) = x\cdot \text{Sigmoid}(x) = x\cdot (1 + \text{TanH}(x/2)/2
$$
$$
\text{GELU}(x) = x\cdot \Phi(x)\approx \frac{x}{2}(1+\text{TanH}(\sqrt{{2}/{\pi}}(x + ax^3)))
$$
where $\Phi$ is Gaussian CDF and $a= 0.044715$.

Here we are concerned about efficient approximation of such activations while preserving the intricate non-linear regions in order to perform complex machine learning tasks accurately.
One solution is to use low-precision inputs, e.g. 16-bit BFloat16 \cite{bfloat, tensorflow}, to such functions in order to achieve high-performance activations (with almost no loss in accuracy). Other approaches involve software optimizations through polynomial approximations \cite{jsfi19}, such as, Pad\'e rational polynomials, piece-wise minimax polynomials, and Taylor expansions. 
These polynomials are typically evaluated using fused-multiply-and-add (FMA) operations via Horner's rule. 

We propose a novel algorithm, K-TanH (Algorithm \ref{alg:approx_tanh}) for approximation of TanH function using only integer operations, such as, shift and add/subtract, \textit{eliminating the need for any multiplication or floating point operations}. This can significantly improve area/power profile for K-TanH. For example, INT8 ADD is $>100\times$ power efficient than Float32 MULT (see \cite{AIAccelaration} for area/power profile for various operations). Integer operations of K-TanH are parameterized where the parameters are encoded in small look-up tables (LUT) that can be fit in computer registers for fast access. Flexible design of K-TanH enables an elegant trade-off among LUT size, accuracy, and throughput. 
High accuracy yet low area/power profile makes K-TanH attractive to deploy in data servers as well as in mobile/embedded devices for both training and inference. 

We assume the numbers are represented in IEEE float format, e.g., $x: (s, E, M) = (-1)^s\cdot 2^E \cdot (1 + M/2^p)$, where $s, E, M, p$ are sign, bias-added exponent,  mantissa, number of mantissa bits, respectively, and all of them are non-negative integers. For float32, bits are allotted as $(s, E, M) = (1, 8, 23)$. BFloat16 $(s, E, M) = (1, 8, 7)$ is a popular 16-bit low-precision numerical format for DNN training/inference. 
K-TanH is compatible with multiple such data formats, although we focus on BFloat16 for the succinctness of discussion.
K-TanH is suitable for efficient hardware design. However, we provide AVX512 implementation for it on General purpose Intel CLX processors for BFloat16 inputs to demonstrate its throughput vis-\`a-vis other approximations (Table \ref{table:compare}). 

Finally, we validate the accuracy of K-TanH experimentally on real-life DL workload.  
We achieve state-of-the-art accuracy and convergence results  training language translation model GNMT \cite{gnmt} for German to English translation on WMT16 data sets via K-TanH on BFloat16 inputs. 
\section{Approximation of TanH}\label{sec:tanh}
Deep Learning models are observed to be resilient to small perturbations. For efficiency of DL workloads, there exist several approximation methods to eliminate the computationally expensive exponentiation of TanH. These methods incur various level of loss of accuracy due to approximation. 
Here we investigate a couple of them: 

1) Piece-wise Minimax approximation that fits a polynomial of degree $n$ on TanH values for an interval of inputs, 2) Rational Pad\'e $_{[p/q]}$ approximation which finds an appropriate ratio between two polynomials of degrees $p$ and $q$ (See \cite{jsfi19}).

\subsection{Minimax Polynomial}
For piece-wise minimax polynomial approximation, we first divide the input range into intervals and then for each interval $[a, b]$ we fit a polynomial $P(x)$ of degree $n$ to minimize
$$
\max_{a\le x \le b}\sabs{\text{TanH}(x) - P(x)}
$$
Here we investigate minimax polynomials of degrees 2 and 3. 

\subsection{Pad\'e rational Polynomial}
Low-degree polynomials may not yield good approximation to TanH as it has two asymptotes. Rational Pad\'e approximation can be a better candidate for this. Pad\'e approximation of some function $f$ is the ratio of two polynomials with degrees $p$ and $q$.
$$
\text{Pad\'e}_{[p/q]f}(x) = \frac{\sum_{i=0}^p a_ix^i}{\sum_{i=0}^q b_ix^i}
$$
Coefficients $\{a_i\}$ and $\{b_i\}$ are calculated as follows. Consider the first $p+q$ derivatives of $f$ at zero and solve the system of equations:
\begin{eqnarray*}
f(0) &=& \text{Pad\'e}_{[p/q]f}(0)
\\
f^{(1)}(0) &=& \text{Pad\'e}^{(1)}_{[p/q]f}(0)
\\
&& \cdot \cdot \cdot
\\
f^{(p+q)}(0) &=& \text{Pad\'e}^{(p+q)}_{[p/q]f}(0)
\end{eqnarray*}

We compare results with $\text{Pad\'e}_{[3/2]}$ and $\text{Pad\'e}_{[7/8]}$. 

\subsection{K-TanH: Our Algorithm to Approximate TanH}

\begin{figure*}[t]
\centering
\includegraphics[scale = 0.25]
{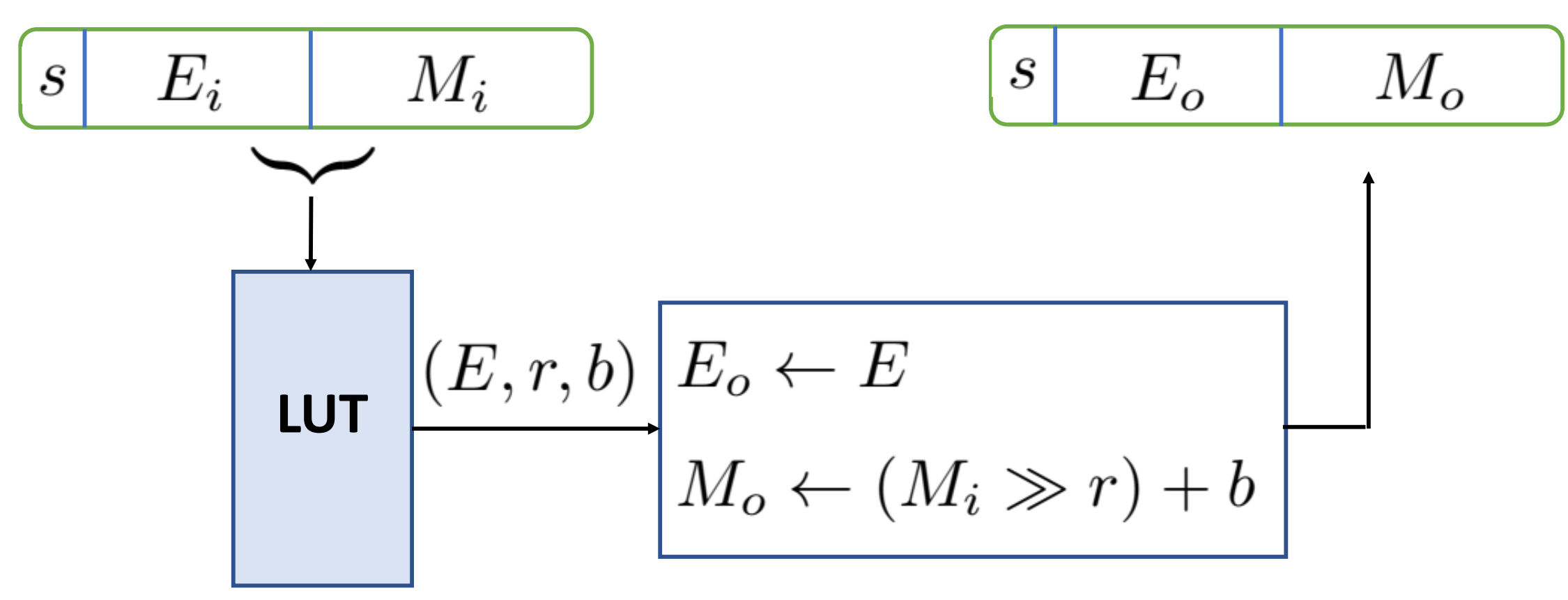}
\caption{Flow chart for K-TanH (Algorithm \ref{alg:approx_tanh}).}
\label{fig:tanhFlowChart}
\end{figure*}

We want to replace expensive TanH with an efficient parametric  transformation function $f$, s.t., $f(x; \theta) \approx \text{TanH}(x)$.
Here choice of $f$ is motivated by efficient low-precision Deep Learning compute, especially, INT8  operations. 
For an input, $x_i = (s_i, e_i, m_i)$, we first determine the interval $t$ for it by evaluating lower bits of exponent and higher bits of mantissa. Then we fetch parameters $\theta_t = (E_t, r_t, b_t)$ for $f$ (stored in LUT), such that, approximate TanH is represented by $y_i = (s_o, E_o, M_o)$, where $s_o=s_i$, $E_o = E_t$, $M_o = (m_i \gg r_t) + b_t$, where $\gg$ denotes right shift. See  Algorithm \ref{alg:approx_tanh}. We set $T_1=0.25$ and $T_2=3.75$, where bit representations in BFloat16 are $T_1:(s, E, M)=(0,01111101,0000000)$ and $T_2:(s, E, M)=(0,10000000,1110000)$. Note that, the conditions in  Lines 3-4 in Algorithm \ref{alg:approx_tanh} can be implemented using INT operations by inspecting $8$ bits of input exponent and $3$ bits of input mantissa. For small magnitude inputs $<0.25$, K-TanH simply bypasses the input to output with no computation involved. Similarly, for large values, it simply returns a fixed value. 

\begin{figure}[t]
\centering
\begin{algorithm}[H]
\caption{K-TanH}\label{alg:approx_tanh}
1. \textbf{Input:} Input $x_{i} = (s_{i}, E_{i}, M_{i})$, 
Parameter Tables $\mathcal{T}_E$, $\mathcal{T}_r$, $\mathcal{T}_b$.\\
2. \textbf{Output:} 
Output $y_{o} = (s_{o}, E_{o}, M_{o})$
\\
3. If $\sabs{x_{i}} < T_1$, 
\quad $y_o \leftarrow x_{i}$, \quad i.e., $(s_o, E_o, M_o) = (s_i, E_i, M_i)$. 
\\
4. Else If $\sabs{x_{i}} > T_2$, 
\quad $y_o \leftarrow s_{i}\cdot 1$, \quad i.e., $(s_o, E_o, M_o) = (s_i, E_{bias}, 0)$.
\\
5. Else, 
\\
6. \quad Form bit string $t$ using lower bits of $E_{i}$ and higher bits of $M_{i}$.
\\
7. \quad Fetch parameters $\theta_t = (E_t, r_t, b_t)$ from $\mathcal{T}_E$, $\mathcal{T}_r$, $\mathcal{T}_b$ using index $t$.
\\
8. \quad $s_o \leftarrow s_{i}, E_o \leftarrow E_t, M_o \leftarrow (M_{i} \gg r_t) + b_t$  
\\
9. Return $y_o$
\end{algorithm} 
\end{figure}

\subsection{Optimizing Parameters for K-TanH}
K-TanH is compatible with various input formats. However, we focus on optimizing BFloat16 inputs only. 
Here is a brief description of how to construct optimized parameters for K-TanH. We consider only the non-negative inputs as TanH is symmetric around zero. We divide the inputs into intervals based on lower bits of exponent and higher bits of mantissa (2 LSBs of exponent and 3 MSBs of mantissa produce 32 intervals). Note that  for such an interval $t$, all the inputs $\{x_i\}$ have common exponent, i.e., 
$x_i: (s, e, m_i) = (-1)^s \cdot 2^{e} \cdot (1 + m_i/2^p)$, where $p$ is number of precision bits for mantissa.
\\\\
\textbf{Step 1}: 
$y_i = \text{TanH}(x_i)
= (-1)^s\cdot 2^{E_i}\cdot (1 + M_i/2^q)$, $q$ is the number of mantissa bits.
\\\\
\textbf{Step 2}: 
For an interval $t$, all $\{y_i\}$ may not have a common exponent (e.g., $x_i \in [0.5, 0.625))$. Therefore, we first transform $\{y_i\}$ to the nearest $\{\hat{y}_i\}$ such that $\{\hat{y}_i\}$ have a common exponent, i.e., 
$$
\hat{y}_i = (-1)^s \cdot 2^E \cdot (1 + \hat{M}_i/2^q)
$$ 
We find $E$ and $\hat{M}_i$ by 
minimizing   
$$
\argmin_{E, \hat{M}_i \in \mathbb{Z}}\sum_i(y_i - \hat{y}_i)^2,
\quad \text{s.t.} \quad E \in \{E_i\}, \hat{M}_i \in [0, 127]
$$
We pick $E$ from the set of exponents $\{E_i\}$. If $E = E_j$, then, $\hat{M}_j = M_j$, for all $j$. If $E > E_j$, then, $\hat{M}_j = 0$. Similarly, for $E < E_j$, $\hat{M}_j = 2^q - 1$.  
Store this $E$ in the parameter table $\mathcal{T}_E$. 
\\\\
\textbf{Step 3}: 
We find optimized shift and add parameters $r$ and $b$, respectively, for the interval of inputs $\{x_i\}$, by solving the following optimization problem. 
\begin{eqnarray}\label{eqn:opt}
\nonumber 
&& \argmin_{r, b \in \mathbb{Z}} 
\sum_i (\hat{M}_i - (m_i/2^r+b))^2, 
\\
\text{ s.t. } 
&&
0 \le r \le r_{max} \le p, 
\quad
b_{min} \le b \le b_{max}
\end{eqnarray}
$b_{min}$ and $b_{max}$ are chosen carefully such that there is no overflow/underflow from INT shift and add operations on mantissa. 
%
Let \textit{mantissa\_idx\_val} be the  decimal value of the $s$ number of MSBs of mantissa used for table indexing; e.g., if 3 MSBs of mantissa 1100111 are used for indexing,  mantissa\_idx\_val = $(110)_2= 6$. We set
%
$$
b_{max} 
= 2^p - 1 - \floor{((\text{ mantissa\_idx\_val}+1)\cdot 2^{p-s}-1)/2^r}
$$
\begin{eqnarray}\label{eqn:b_min}
b_{min} 
&=& -1\cdot   \text{mantissa\_idx\_val} \cdot \floor{2^{p-s-r}}
\end{eqnarray}
For fixed $r$, we find $b$ solving a least square problem (then round it). If $b < b_{min}$, $b = b_{min}$ and if $b > b_{max}$, $b = b_{max}$.
Store optimized $r$ and $b$ in $\mathcal{T}_r$, $\mathcal{T}_b$, respectively.

Finding the optimized tables is  one time offline compute process. Also, we want to fit each table in a register of a general purpose machine for quick access. E.g., to fit each table in a 512-bit register for Intel AVX512 SIMD instructions, we use 5-bit indexing (2 LSBs of exponent and 3 MSBs of mantissa) to create 32 entries (32 intervals of the input magnitude), each holding up to 16 bit integer values. Our parameter values are 8-bit only, so we can create 64 intervals to achieve more accurate approximation. However, experimentally, 32 entries suffices.

\begin{table*}[t]
\centering
\caption{Optimized parameter table for BFloat16 inputs in Algorithm \ref{alg:approx_tanh}, where $t$ is created using 2 LSBs of $E_i$ and 3 MSBs of $M_i$. Each table $\mathcal{T}_E$, $\mathcal{T}_r$, $\mathcal{T}_b$ can be fit in one 512-bit register. 
}
\begin{tabular}{c|ccc||c|ccc}
    Index $t$ 
    & $E_t$ & $r_t$ & $b_t$ & Index $t$ & $E_t$ & $r_t$ & $b_t$ \\
    \hline
    00111 & 126 & 6 & 126 & 10111 & 126 & 1 & 4 \\ 
    00110 & 126 & 6 & 126 & 10110 & 126 & 1 & 4 \\
    00101 & 126 & 6 & 126 & 10101 & 126 & 1 & 4 \\
    00100 & 126 & 6 & 126 & 10100 & 126 & 1 & 3 \\
    00011 & 126 & 4 & 123 & 10011 & 126 & 1 & 2 \\
    00010 & 126 & 4 & 123 & 10010 & 126 & 1 & -1 \\
    00001 & 126 & 4 & 122 & 10001 & 126 & 1 & -4 \\
    00000 & 126 & 2 & 119 & 10000 & 125 & 0 & 112 \\
    11111 & 126 & 4 & 110 & 01111 & 125 & 0 & -18 \\
    11110 & 126 & 2 & 89 & 01110 & 125 & 0 & -15 \\
    11101 & 126 & 2 & 89 & 01101 & 125 & 0 & -12 \\
    11100 & 126 & 2 & 88 & 01100 & 125 & 0 & -10 \\
    11011 & 126 & 1 & 73 & 01011 & 125 & 0 & -7 \\
    11010 & 126 & 1 & 73 & 01010 & 125 & 0 & -6 \\
    11001 & 126 & 1 & 72 & 01001 & 125 & 0 & -4 \\
    11000 & 126 & 0 & 65 & 01000 & 125 & 1 & 1 \\
    \hline
  \end{tabular}
\label{table:param_bfloat16}
\end{table*}

\subsection{Performance of K-TanH}

\begin{figure*}[!t]
\centering
\includegraphics[scale = 0.5]
{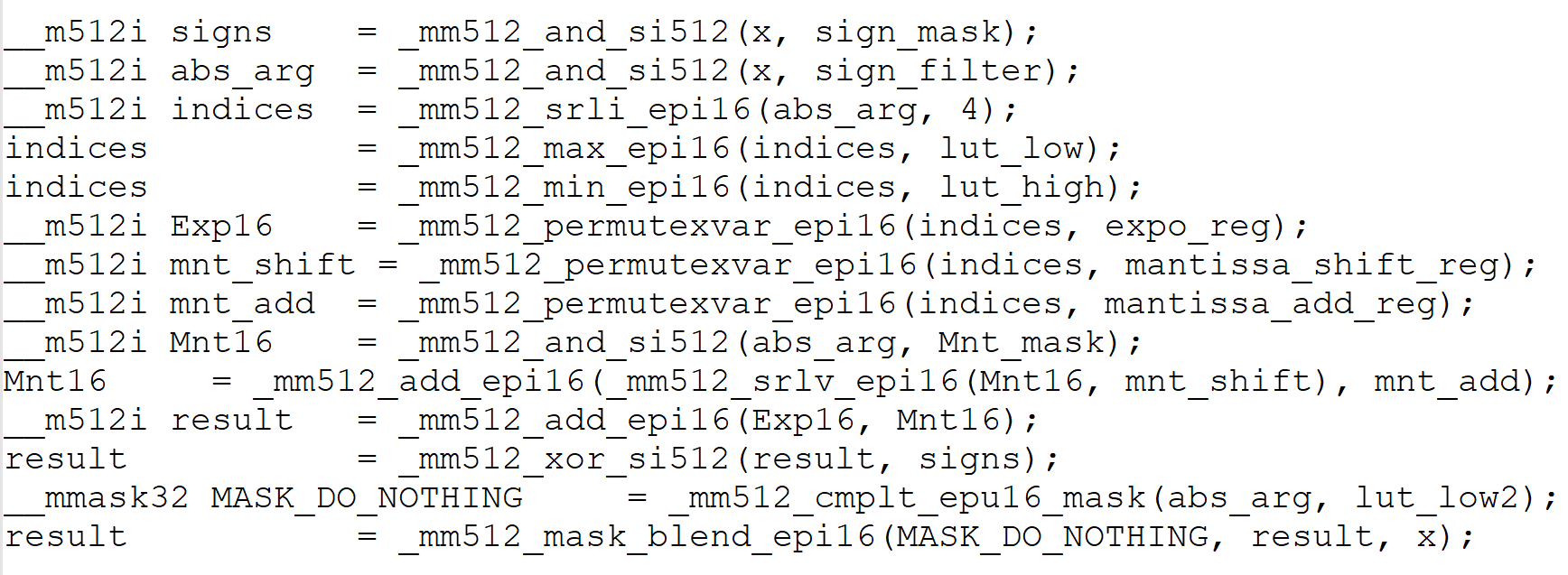}
\caption{AVX512 Intrinsics for Algorithm \ref{alg:approx_tanh}}
\label{fig:Intrinsics}
\end{figure*}

We assume BFloat16 input for K-TanH (Float32 to BFloat16 conversion cost is not considered). 
Software performance of our implementation of K-TanH is based on and limited by available AVX512 instructions. For example, potential INT8 operations of K-TanH are implemented as INT16 ops due to lack of INT8 shift and INT8 concatenation. Table \ref{table:compare} shows performance of various TanH approximations. K-TanH has the potential to deliver $<0.1$ cycles/TanH (i.e. $>15\times$ speed up over Intel SVML) with appropriate hardware support. Figure \ref{fig:Intrinsics} shows the AVX512 Intrinsics for K-TanH.
\begin{table*}[t]
\centering
\caption{AVX512 implementation of various approximation of TanH (Intel CLX processors)}
\begin{tabular}{c||cc|c|c|c}
Approx Alg  
    \multirow{2}{*}{} & Max Err & Rel Err & Cycles & Speed & float\\
& ($\times 10^{-2}$) & (\%) & per TanH & Up & ops\\
    \hline
Intel SVML (high prec) & $-$ & $-$ & $1.53$ & $1$ $\times$ & \checkmark \\
Intel SVML (low prec) & $-$ & $-$ & $0.95$ & $1.61$ $\times$  & \checkmark \\
Rational Pad\'e 7/8 & $0.01$ & $0.01$ & $0.59$ & $2.59$ $\times$ & \checkmark \\
Taylor approx degree 3 & $0.04$ & $28.57$ & $0.47$ & $3.26$ $\times$ & \checkmark \\
Taylor approx degree 2 & $0.43$ & $28.57$ & $0.42$ & $3.64$ $\times$ & \checkmark \\
Minimax poly degree 3 & $0.01$ & $0.01$ & $0.42$ & $3.64$ $\times$ & \checkmark \\
Rational Pad\'e 3/2 & $2.35$ & $2.59$ & $0.39$ & $3.92$ $\times$ & \checkmark \\
Minimax poly degree 2 & $0.07$ & $8.36$ & $0.35$ & $4.37$ $\times$ & \checkmark \\
\hline 
\textbf{K-TanH (BFloat16)} & $1.67$ & $3.03$ & \textbf{0.28} & \textbf{5.46} $\times$ & \textbf{$\times$} \\
    \hline
  \end{tabular}
\label{table:compare}
\end{table*}
%
%
%
%
\section{Experiments}\label{sec:experiments}
%
We test the accuracy of K-TanH (Algorithm \ref{alg:approx_tanh}) on complex real life problems, such as, language translation. Specifically, we train Google's 8-layer Neural Machine Translation (GNMT) \cite{gnmt} on WMT16 German to English (De-En) data sets for 5 epochs. 
K-TanH consistently improved on SVML baseline Bleu score despite being the most efficient (Tables \ref{table:De_En_train},\ref{table:De_En_test}).

\begin{table*}[t]
\centering
\caption{De-En: Best Bleu scores (higher the better) of 8-layer GNMT on training data for various approximation of TanH activation.}
  \begin{tabular}{c| c c c c c c }
Epoch & SVML
& \textbf{K-TanH} & Minimax 3 & Minimax 2 & Pad\'e 7/8 &  Pad\'e 3/2    \\
    \hline
1 & 18.6 & \textbf{19.7} & 19.1 & 18.0 & 18.4 & 18.2 \\
2 & 22.7 & \textbf{23.0} & 22.9 & 22.4 & 22.7 & 22.7 \\
3 & 23.8 & \textbf{24.3} & 24.2 & 23.8 & 24.5 & 24.0 \\
4 & 25.3 & \textbf{25.4}  & 25.3 & 25.1 & 25.1 & 25.3 \\
5 & 26.1 & \textbf{26.5}  & 26.2 & 26.0 & 26.1 & 25.9 \\
\hline
\end{tabular}
 \label{table:De_En_train}
\end{table*}
\begin{table*}[t]
\centering
\caption{De-En: Best Bleu scores (higher the better) of 8-layer GNMT on test data for various approximation of TanH activation.}
  \begin{tabular}{c| c c c c c c }
Epoch & SVML
& \textbf{K-TanH} & Minimax 3 & Minimax 2 & Pad\'e 7/8 &  Pad\'e 3/2    \\
    \hline
1 & 18.9 & \textbf{20.1} & 19.7 & 18.2 & 18.4 & 18.5 \\
2 & 23.5 & \textbf{23.7} & 23.3 & 22.9 & 23.5 & 23.2 \\
3 & 24.6 & \textbf{25.1} & 24.9 & 24.2 & 25.0 & 24.5 \\
4 & 26.0 & \textbf{26.3}  & 26.1 & 25.2 & 26.3 & 26.0 \\
5 & 26.7 & \textbf{26.8}  & 26.8 & 26.3 & 26.9 & 26.9 \\
\hline
\end{tabular}
 \label{table:De_En_test}
\end{table*}

\subsection{Ablation study} 
We investigate the behavior of K-TanH on sub-optimal parameters. For this, 
we solve (\ref{eqn:opt}) by setting $b_{min} = 0$ in (\ref{eqn:b_min}). For K-TanH using such sub-optimal parameters, training and test Bleu scores (after 5 epochs) drop by  $>2.6\%$ and $>3.3\%$ from baseline, respectively. This indicates the importance of optimizing the parameters as in (\ref{eqn:opt}) with appropriately chosen constraints. 


Overall, our K-TanH is  hardware-friendly, efficient and accurate approximation of TanH, and it is superior to existing competitive approximation schemes while achieving state-of-the-art results on a challenging DL workload.


\newpage
\bibliographystyle{plainnat}

\end{document}